\documentclass[letterpaper]{article} 
\usepackage{aaai25}  
\usepackage{times}  
\usepackage{helvet}  
\usepackage{courier}  
\usepackage[hyphens]{url}  
\usepackage{graphicx} 
\usepackage{natbib}  
\usepackage{caption} 
\usepackage{algorithm}
\usepackage{algorithmic}

\usepackage{multirow}
\usepackage{makecell}
\usepackage{booktabs}
\usepackage{amssymb}
\usepackage{pifont}
\newcommand{\cmark}{\ding{51}}%
\newcommand{\xmark}{\ding{55}}%
\usepackage{newfloat}
\usepackage{listings}
\DeclareCaptionStyle{ruled}{labelfont=normalfont,labelsep=colon,strut=off} 
\floatstyle{ruled}
\newfloat{listing}{tb}{lst}{}
\floatname{listing}{Listing}
\newcommand\blfootnote[1]{
  \begingroup
  \renewcommand\thefootnote{}\footnote{#1}
  \addtocounter{footnote}{-1}
  \endgroup
}

\title{Disentangled Motion Modeling for Video Frame Interpolation}
\author{
    Jaihyun Lew\textsuperscript{\rm 1},
    Jooyoung Choi\textsuperscript{\rm 2},
    Chaehun Shin\textsuperscript{\rm 2},
    Dahuin Jung\textsuperscript{\rm 3,\dag},
    Sungroh Yoon\textsuperscript{\rm 1,2,4,\dag}
}
\affiliations{
    \textsuperscript{\rm 1}Interdisciplinary Program in AI, Seoul National University\\
    \textsuperscript{\rm 2}Department of Electrical and Computer Engineering, Seoul National University\\
    \textsuperscript{\rm 3}School of Computer Science and Engineering, Soongsil University \\
    \textsuperscript{\rm 4}AIIS, ASRI and INMC, Seoul National University\\
    \{fudojhl, jy\_choi, chaehuny\}@snu.ac.kr, dahuin.jung@ssu.ac.kr, sryoon@snu.ac.kr
}
\begin{document}

\twocolumn[{
\renewcommand\twocolumn[1][]{#1}
\maketitle
\vspace{-1em}
\begin{center}
\centering
    \centering
    \includegraphics[width=\textwidth]{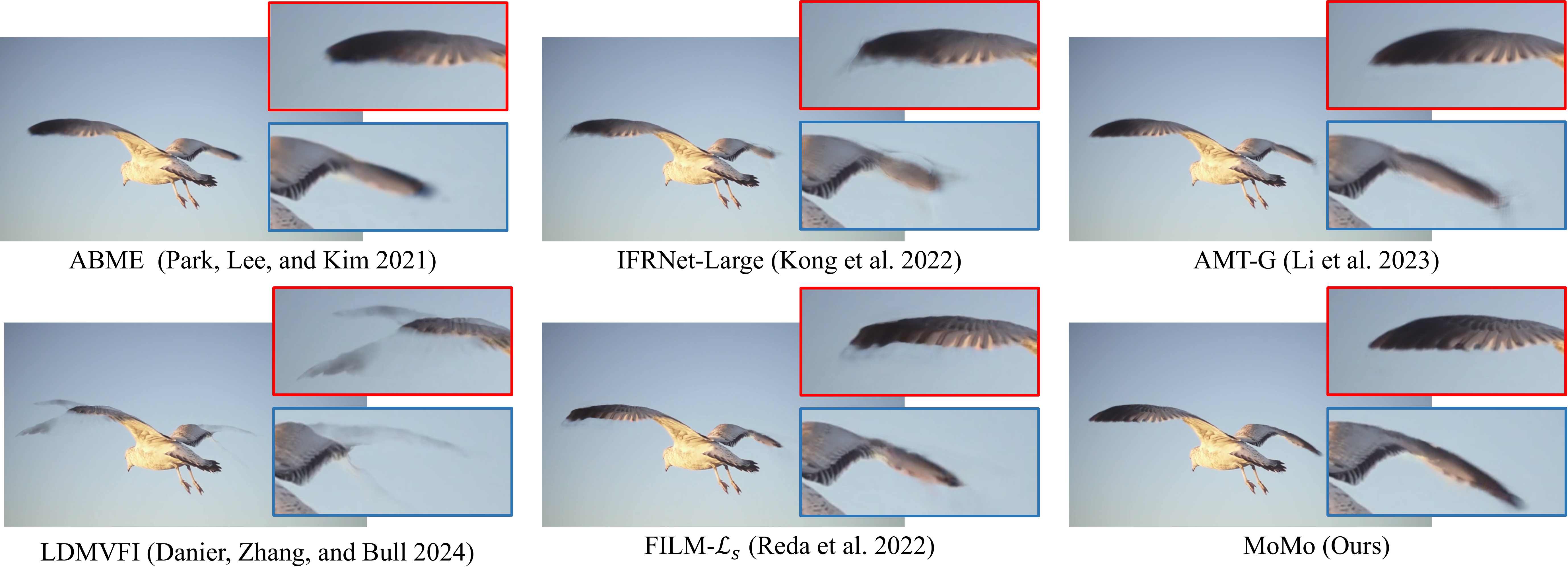}
    \vspace{-2em}
    \captionof{figure}{Video frame interpolation results of our proposed method called MoMo with comparison to state-of-the-art methods. MoMo produces the most visually pleasant result, owing to proper modeling of the intermediate motion.}
    \label{fig:main}
\end{center}
}]

\blfootnote{$^\dagger$ Corresponding authors}
\blfootnote{Code available at: \url{https://github.com/JHLew/MoMo}}

\begin{abstract}
Video Frame Interpolation (VFI) aims to synthesize intermediate frames between existing frames to enhance visual smoothness and quality.
Beyond the conventional methods based on the reconstruction loss, recent works have employed generative models for improved perceptual quality. 
However, they require complex training and large computational costs for pixel space modeling.
In this paper, we introduce disentangled Motion Modeling (MoMo), a diffusion-based approach for VFI that enhances visual quality by focusing on intermediate motion modeling.
We propose a disentangled two-stage training process.
In the initial stage, frame synthesis and flow models are trained to generate accurate frames and flows optimal for synthesis.
In the subsequent stage, we introduce a motion diffusion model, which incorporates our novel U-Net architecture specifically designed for optical flow, to generate bi-directional flows between frames.
By learning the simpler low-frequency representation of motions, MoMo achieves superior perceptual quality with reduced computational demands compared to the generative modeling methods on the pixel space.
MoMo surpasses state-of-the-art methods in perceptual metrics across various benchmarks, demonstrating its efficacy and efficiency in VFI.
\end{abstract}

\section{Introduction}
Video Frame Interpolation (VFI) is a crucial task in computer vision that aims to synthesize absent frames between existing ones in a video.
It has a wide spectrum of applications, such as slow motion generation~\cite{superslomo}, video compression~\cite{wu2018video}, and animation production~\cite{AnimeInterp}.
Its ultimate goal is to elevate the visual quality of videos through enhanced motion smoothness and image sharpness.
Motions, represented by optical flows~\cite{pwcnet, raft}
and realized by warping, have been central to VFI's development as recent innovations in VFI have mostly been accomplished along with advances in intermediate motion estimation~\cite{qvi, allatonce, bmbc, abme, xvfi, film, uprnet}.

However, these approaches often result in perceptually unsatisfying outcomes due to their reliance on $L_1$ or $L_2$ objectives, leading to high PSNR scores yet poor perceptual quality~\cite{srgan, lpips}.
To address this matter, recent advancements~\cite{cain, superslomo, softsplat, ImprovingAnimeInterp} have explored the use of deep feature spaces to achieve improved quality in terms of human perception~\cite{perceptualloss, lpips}.
Additionally, the integration of generative models into VFI~\cite{LDMVFI, pervfi} has introduced novel pathways for improving the visual quality of videos but has primarily focused on modeling pixels or latent spaces directly, which demands high computational resources.

We introduce disentangled \textbf{Mo}tion \textbf{Mo}deling (MoMo), a perception-oriented approach for VFI, focusing on the modeling of intermediate motions rather than direct pixel generation.
Here, we employ a diffusion model~\cite{ddpm} to generate bi-directional optical flow maps, marking the first use of generative modeling for motion in VFI. 
We propose to disentangle the training of frame synthesis and intermediate motion prediction into a two-stage process:
the initial stage includes the training of a \emph{frame synthesis model} and fine-tuning of an \emph{optical flow model}~\cite{raft}.
The frame synthesis model is designed to correctly synthesize an RGB frame given a pair of frames and their corresponding flow maps.
In the subsequent stage of training, we train our \emph{motion diffusion model}, which generates the intermediate motions the frame synthesis model uses to create the final interpolated frame during inference. In this stage of training, the optical flow model fine-tuned in the first stage serves as a teacher to provide pseudo-labels for the motion diffusion model.
We also propose a novel architecture for our motion diffusion model, inspired by the nature of optical flows, enhancing both computational efficiency and performance.

Our experiments validate the effectiveness and efficiency of our proposed training scheme and architecture, demonstrating superior performance across various benchmarks in terms of perceptual metrics,
with approximately $70 \times$ faster runtime compared to the existing diffusion-based VFI method~\cite{LDMVFI}.
By prioritizing the generative modeling of motions, our approach enhances visual quality, effectively addressing the core objective of VFI.

Our contributions can be summarized as follows:
\begin{itemize}
    \item  We introduce MoMo, a diffusion-based method focusing on generative modeling of bi-directional optical flows for the first time in VFI.
    \item We propose to disentangle the training of frame synthesis and intermediate motion modeling into a two-stage process, which are the crucial components in VFI.
    \item We introduce a novel diffusion model architecture suitable for optical flow modeling, boosting efficiency and quality.
\end{itemize}

\section{Related Work}
\subsection{Flow-based Video Frame Interpolation}
In deep learning-based Video Frame Interpolation (VFI), optical flow-based methods have recently become prominent,
typically following a common two-stage process.
First, the flows \emph{to} or \emph{from} the target intermediate frame is estimated, which involves warping of the input frame pair with the estimated flows. Then, a synthesis network merges the warped frames to produce the final frame.
Recent advances in VFI quality have progressed with enhancements in intermediate flow predictions~\cite{rife, ifrnet, vfiformer, emavfi, amt}, sparking specialized architectures to improve flow accuracy~\cite{bmbc, abme, biformer}.
Following this direction of studies, our work aims to focus on improving intermediate flow prediction.
Unlike most methods that heavily rely on reconstruction loss for end-to-end training, with optional flow distillation loss~\cite{rife, ifrnet} for stabilized training, our approach employs disentangled and direct supervision solely on flow estimation, marking an innovation in VFI research.

\subsection{Perception-oriented Restoration}
Conventional restoration methods in computer vision, including VFI, focused on minimizing $L_1$ or $L_2$ distances, often resulting in blurry images~\cite{srgan} due to prioritizing of pixel accuracy over human visual perception. Recent studies have shifted towards deep feature spaces for reconstruction loss~\cite{perceptualloss} and evaluation metrics~\cite{lpips, dists}, demonstrating that these align better with human judgment. These approaches emphasize perceptual similarities over traditional metrics like PSNR, signaling a move towards more visually appealing, photo-realistic image synthesis.

Ever since the pioneering work of SRGAN~\cite{srgan}, generative models have been actively used to enhance visual quality in restoration tasks~\cite{sr3, pulse}.
The adoption of generative models has also been explored in VFI~\cite{mcvd, LDMVFI, pervfi}. LDMVFI~\cite{LDMVFI}, closely related to our work, use latent diffusion models~\cite{ldm} to enhance perceptual quality.
Our method aligns with such innovations but uniquely focuses on generating optical flow maps, differing from prior generative approaches that directly model the pixel space.

\subsection{Diffusion Models}
Diffusion models~\cite{sohl2015deep, ddpm} are popular generative models that consist of forward and reverse process. Initially, the forward process incrementally adds noise to the data $\textbf{x}_0$ over $T$ steps via a predefined Markov chain, resulting in $\textbf{x}_T$ that approximates a Gaussian noise. The diffused data $\textbf{x}_t$ is obtained through forward process:
\begin{equation}
\label{eq:forward}
    \textbf{x}_t=\sqrt{\alpha_{t}} \textbf{x}_0 + \sqrt{1-\alpha_{t}}\epsilon,
\end{equation}
where $\alpha_{t}\in\{\alpha_{1}, ..., \alpha_{T}\}$ is a pre-defined noise schedule.
Then, the reverse process undoes the forward process by starting from Gaussian noise $\textbf{x}_T$ and gradually denoising back to $\textbf{x}_0$ over $T$ steps. Diffusion models train a neural network to perform denoising at each step, by minimizing the following objective:
\begin{equation}
    L=\mathbb{E}_{\textbf{x}_0, \epsilon \sim \mathcal{N}(\mathbf{0}, \mathbf{I}), t \sim \mathcal{U}(1, T)}\left\|{\epsilon}-{\epsilon}_\theta\left({\textbf{x}}_t, t\right)\right\|_2.
\end{equation}
While a commonly used approach is to predict the noise as above ($\epsilon$-prediction), there are some alternatives, such as $\textbf{x}_0$-prediction~\cite{dalle2} which predicts the data $\textbf{x}_0$ itself or $\textbf{v}$-prediction~\cite{progressivedistill}, beneficial for numerical stability.

\begin{figure*}[t!]
    \centering
    \includegraphics[width=\textwidth]{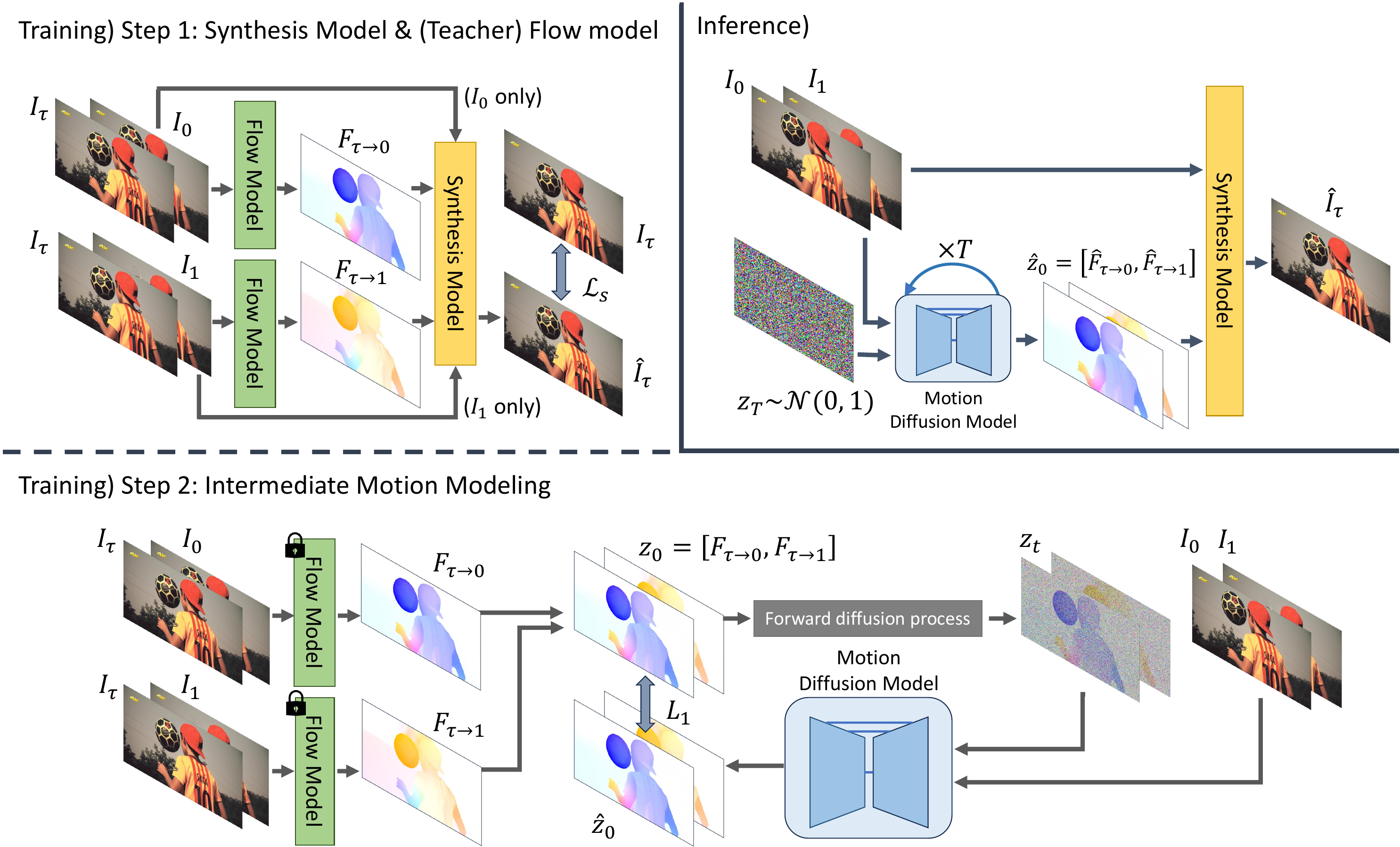}
    \caption{
    Overview of our entire framework.
    The training procedure operates in two stages. Initially, we train a frame synthesis network and an optical flow model, with the latter providing pseudo-labels for the second stage. In the second stage of training, we focus on training a Motion Diffusion Model to predict bi-directional flow between frames. During inference, the Motion Diffusion Model generates flow fields given the input frame pair, which the frame synthesis model uses to generate the output.
    }
    \vspace{-10pt}
    \label{fig:overview}
\end{figure*}

Diffusion model synthesizes the data in an iterative manner following the backward process, resulting in high perceptual quality of image samples or video samples~\cite{imagen, imagenvideo}.
Further, we are motivated by optical flow modeling with diffusion models in other tasks~\cite{ddvm, lfdm}, and aim to leverage the benefit of diffusion models for optical flow synthesis in video frame interpolation.
Although an existing work employs diffusion model for video frame interpolation task~\cite{LDMVFI}, our method synthesizes the intermediate optical flows rather than directly synthesizing the RGB frames.

\section{Method}
\subsection{Overview}
In this paper, we focus on the goal of synthesizing an intermediate frame $I_\tau$ between consecutive frames $I_0$ and $I_1$, where $0 < \tau < 1$.
Our method adopts a two-stage training scheme to disentangle the training of motion modeling and frame synthesis (Fig.~\ref{fig:overview}). In the first stage, we train a frame synthesis network to synthesize an RGB frame from neighboring frames and their bi-directional flows. Then, we fine-tune the optical flow model to enhance flow quality. In the second stage, the fine-tuned flow model serves as a teacher for training the motion diffusion model. During inference, this motion diffusion model generates intermediate motion (bi-directional flow maps in specific), which the synthesis network uses to produce the final RGB frame.

\subsection{Synthesis and Teacher Flow Models}
\label{sec:synth_and_teacher}
We propose a synthesis network $\mathcal{S}$, designed to accurately generate an intermediate target frame using a pair of input frames and their corresponding optical flows from the target frames.
Specifically, given a frame pair of $I_0, I_1$, and the target intermediate frame $I_\tau$, we first use an optical flow model $\mathcal{F}$ to obtain the bi-directional flow from the target frame to the input frames:
\begin{equation}
    F_{\tau \rightarrow i} = \mathcal{F}(I_\tau, I_i), i \in \{0, 1\},
\end{equation}
where $i$ denotes the index of input frames.
With the estimated flows and their corresponding frames from the target frame, we synthesize $\hat{I}_\tau$, which aims to recover the target frame $I_\tau$.
\begin{equation}
    \hat{I}_\tau = \mathcal{S}(I_{in}, F_\tau), 
\end{equation}
where $I_{in}$ denotes the input frame pair $\{I_0, I_1 \}$ and $F_\tau$ denotes the corresponding flow pair $\{F_{\tau\rightarrow0}, F_{\tau\rightarrow1}\}$.

\begin{figure*}[t]
    \centering
    \includegraphics[width=\textwidth]{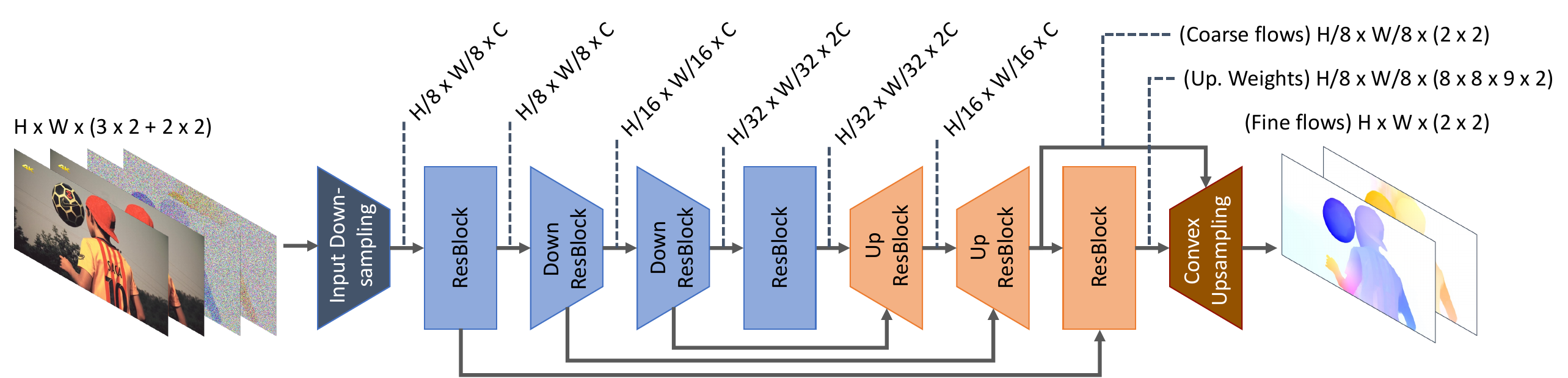}
    \caption{Architecture of our motion diffusion model. The input pair frames are downsampled to an $8\times$ smaller size and goes through a 3-level U-Net, which outputs a pair of coarse flow maps and their corresponding weight masks for upsampling. The convex upsampling layer takes the coarse flow maps and weight masks to return the full resolution flow maps.}
    \vspace{-7pt}
    \label{fig:diff_arch}
\end{figure*}

We adopt pre-trained RAFT~\cite{raft} for optical flow model $\mathcal{F}$, and train the synthesis network $\mathcal{S}$ from scratch.
We use an alternating optimization of two models $\mathcal{S}$ and $\mathcal{F}$.
We first fix $\mathcal{F}$ to the pre-trained state, and train the synthesis network $\mathcal{S}$.
Once the training of $\mathcal{S}$ converges, we freeze $\mathcal{S}$, and fine-tune $\mathcal{F}$.
We fine-tune $\mathcal{F}$ so that it could provide better estimations as the teacher in the next stage of training.
Note that the flow model $\mathcal{F}$ is not used during inference, but serves its purpose as the teacher for intermediate motion modeling described in Sec.~\ref{sec:motion_modeling}.

\paragraph{Objective}
\label{sec:objective}

For optimization, we compute loss on the final synthesized output $\hat{I}_\tau$, with 
a combination of three terms.
First, we use the pixel reconstruction error between the synthesized frame and the target frame: $\mathcal{L}_1 = || I_\tau - \hat{I}_\tau||_1$.
Following recent efforts~\cite{LDMVFI, ImprovingAnimeInterp}, we adopt the LPIPS-based perceptual reconstruction loss $\mathcal{L}_p$~\cite{lpips},
and also exploit the style loss $\mathcal{L}_G$~\cite{gatysstyle} , as its effectiveness has been proved in a recent work~\cite{film}.
By combining the three loss terms, we define our perception-oriented reconstruction loss $\mathcal{L}_s$ for high quality synthesis:
\begin{equation}
    \mathcal{L}_s = \lambda_{1}\mathcal{L}_1 + \lambda_{p}\mathcal{L}_p + \lambda_{G}\mathcal{L}_{G}.
\end{equation}

Further details on our perception-oriented reconstruction loss can be found in the Appendix.

\subsubsection{Recurrent Synthesis}
We build our synthesis network $\mathcal{S}$ to be of recurrent structure, motivated by the recent trend in video frame interpolation~\cite{xvfi, uprnet, film}, due to its great efficiency.
The inputs $I_{in}$ and $F_{\tau}$ are resized to various scales of lower-resolution, and by applying our synthesis module $\mathcal{G}$ recurrently from low-resolution and to higher resolutions, the output frame is synthesized in a coarse-to-fine manner.
For our synthesis module $\mathcal{G}$, we use a simple 3-level hierarchy U-Net~\cite{unet}.
Details on our recurrent synthesis procedure are described in the Appendix.

\subsection{Intermediate Motion Modeling with Diffusion}
\label{sec:motion_modeling}
With a synthesis network fixed,
we focus on modeling the intermediate motions for VFI.
We use our fine-tuned flow model $\mathcal{F}$ as the teacher to train our motion diffusion model $\mathcal{M}$, which generates bi-directional optical flows from the pair of input frames $I_0$ and $I_1$. Denoting concatenated flows $z_0 = \{F_{\tau\rightarrow0}, F_{\tau\rightarrow1}\}$, we train a diffusion model $\mathcal{M}$ by minimizing the following objective:
\begin{equation}
    \mathcal{L}_m = \mathbb{E}_{z_0, t \sim \mathcal{U}(1, T)}[||z_0 - \mathcal{M}(z_t, t, I_0, I_1) ||_1],
\end{equation}
where $z_t$ represents noisy flows diffused by Eq. \ref{eq:forward}. We concatenate $I_0$ and $I_1$ to $z_t$ and keep the teacher $\mathcal{F}$ frozen during training of $\mathcal{M}$. While $\epsilon$-prediction~\cite{ddpm} and $L_2$ norm are popular choices for training diffusion-based image generative models, we found $\textbf{x}_0$-prediction and $L_1$ norm to be beneficial for modeling flows. While image diffusion models utilize a U-Net architecture
that employs input and output of the same resolution for noisy images
that operates fully on the entire resolution,
we introduce a new architecture for $\mathcal{M}$ aimed at learning optical flows and enhancing efficiency, which will be described in the following paragraph.

\subsubsection{Architecture}
An overview of our proposed motion diffusion model architecture is provided at Fig.~\ref{fig:diff_arch}.
In our novel diffusion model architecture designed for motion modeling, we begin by excluding attention layers,
as we have found doing so saves memory without deterioration in performance.
Observing that optical flow maps—our primary target—are sparse representations encoding low-frequency information, we opt to avoid the unnecessary complexity of full-resolution estimation.
Consequently, we initially predict flows at $1/8$ of the input resolution and upsample them by $8 \times$, a method mirroring the coarse-to-fine strategies \cite{raft, gmflow, flowformer}, thus sidestepping the need for full-resolution flow estimation.
We realize this by introducing \textit{input downsampling} and \textit{convex upsampling}~\cite{raft} into U-Net, making our architecture computationally efficient and well-suited to meet our resolution-specific needs.
We elaborate them in the following paragraphs.

\paragraph{Input Downsampling}
Given an input $\{z_t,I_0,I_1\}$ of 10 channels, we downsample and encode it to $1/8$ resolution. Rather than using a single layer to directly apply on the 10 channel input, we separately apply layers $\mathcal{D}_I$ and $\mathcal{D}_z$ on the frames and the noisy flows, respectively:
\begin{equation}
    I_0' = \mathcal{D}_I(I_0),~
    I_1' = \mathcal{D}_I(I_1),~ 
    z_t' = \mathcal{D}_z(z_t).
\end{equation}
By sharing the parameters applied to $I_0$ and $I_1$, we could save the number of parameters required for the downsampling process,
and make it invariant to the order of the two input frames. 
Once we obtain the downsampled features $I_0', I_1', z_t'$, we concatenate and project them to features:
\begin{equation}
    p_t = \mathcal{D}_{p}([I_0', I_1', z_t']),
\end{equation}
where the projection layer $\mathcal{D}_{p}$ is implemented by a single $1\times1$ convolutional layer.

The projected features $p_t$ is then given to a 3-level diffusion U-Net, which produces two outputs: a coarse estimation of flow maps and their corresponding upsampling weight masks. The two outputs are combined in the convex upsampling layer to obtain the final full-scale flow maps.

\paragraph{Convex Upsampling}
Given the coarse flow maps and their corresponding upsampling weight masks, both of size $H/8 \times W/8$,
we attempt to upsample the coarse flows to the original $H \times W$ resolution using a weighted combination of $3 \times 3$ grid of each coarse flow neighbors, by integrating the convex upsampling layer~\cite{raft} to our architecture.
Using the predicted upsampling weight masks of $8 \times 8 \times 9$ channels, we apply softmax on the weights of 9 neighboring pixels, and perform weighted summation with coarse flows to obtain the final upsampled flow map.
An illustrated description is provided in the Appendix. 

This method aligns with $\textbf{x}_0$-prediction, promoting local correlations and differing from $\epsilon$-predictions which necessitate locally independent estimations.
By operating at a reduced resolution, specifically at $64\times$ smaller space, we achieve significant computation savings.
Note that this upsampling layer does not involve any learnable parameters.

\section{Experiments}
\subsection{Experiment Settings}
\subsubsection{Implementation Details}
\label{sec:impl_details}
We train our model on the Vimeo90k dataset~\cite{toflow}, using random $256 \times 256$ crops with augmentations like 90$^\circ$ rotation, flipping, and frame order reversing.
We recommend the reader to refer to the Appendix for further details.

\subsubsection{Evaluation Protocol}
We evaluate on well-known VFI benchmarks: Vimeo90k~\cite{toflow}, SNU-FILM~\cite{cain}, Middlebury (others-set)~\cite{middlebury}, and Xiph~\cite{xiph, softsplat}, chosen for their broad motion diversity and magnitudes.
Following practices in generative models-based restoration~\cite{LDMVFI, liang2022details}, we focus on perceptual similarity metrics LPIPS~\cite{lpips} and DISTS~\cite{dists}, which highly correlates with human perception, for evaluation. While PSNR and SSIM are popular metrics, they have been known to differ from human perception in some aspects, sensitive to imperceptible differences in pixels and preferring blurry samples~\cite{lpips}. 
The full results can be found in the Appendix.
\begin{table*}[t!]
    \centering
    \setlength{\tabcolsep}{1mm}
    \begin{tabular}{l c cc cc cc cc}
    \toprule
        \multirowcell{2}{Method} & \multirowcell{2}{Perception-\\oriented loss} & \multicolumn{2}{c}{FILM-easy} & \multicolumn{2}{c}{FILM-medium} & \multicolumn{2}{c}{FILM-hard} & \multicolumn{2}{c}{FILM-extreme}\\
        && LPIPS & DISTS
        & LPIPS & DISTS
        & LPIPS & DISTS
        & LPIPS & DISTS\\
    \midrule\midrule
        ABME~\cite{abme} 
        & \xmark
        & 0.0222 & 0.0229 
        & 0.0372 & 0.0344 
        & 0.0658 & 0.0496 
        & 0.1258 & 0.0747
        \\
        XVFI$_v$~\cite{xvfi}
        & \xmark
        & 0.0175 & 0.0181
        & 0.0322 & 0.0276
        & 0.0629 & 0.0414
        & 0.1257 & 0.0673
        \\
        IFRNet-Large~\cite{ifrnet} 
        & \xmark
        & 0.0203 & 0.0211
        & 0.0321 & 0.0288
        & 0.0562 & 0.0403
        & 0.1131 & 0.0638
        \\
        RIFE~\cite{rife} 
        & \xmark
        & 0.0181 & 0.0195
        & 0.0317 & 0.0289
        & 0.0657 & 0.0443
        & 0.1390 & 0.0764
        \\
        FILM-$\mathcal{L}_1$~\cite{film} 
        & \xmark
        & 0.0184 & 0.0217
        & 0.0315 & 0.0316
        & 0.0568 & 0.0441
        & 0.1060 & 0.0632
        \\
        AMT-G~\cite{amt} 
        & \xmark
        & 0.0325 & 0.0312 
        & 0.0447 & 0.0395
        & 0.0680 & 0.0506
        & 0.1128 & 0.0686
        \\
        EMA-VFI~\cite{emavfi} 
        & \xmark
        & 0.0186 & 0.0204
        & 0.0325 & 0.0318
        & 0.0579 & 0.0457
        & 0.1099 & 0.0671
        \\ 
        UPRNet-LARGE~\cite{uprnet} 
        & \xmark
        & 0.0182 & 0.0203
        & 0.0334 & 0.0327
        & 0.0612 & 0.0475
        & 0.1109 & 0.0672
        \\
        CAIN~\cite{cain}
        & \cmark
        & 0.0197 & 0.0229  
        & 0.0375 & 0.0347
        & 0.0885 & 0.0606
        & 0.1790 & 0.1042
        \\
        FILM-$\mathcal{L}_{vgg}$~\cite{film}
        & \cmark
        & 0.0123 & 0.0128
        & 0.0219 & 0.0183
        & 0.0443 & 0.0282
        & 0.0917 & 0.0471
        \\
        FILM-$\mathcal{L}_s$~\cite{film}
        & \cmark
        & \underline{0.0120} & \underline{0.0124}
        & \underline{0.0213} & \underline{0.0177}
        & \underline{0.0429} & \underline{0.0268}
        & \underline{0.0889} & \underline{0.0448}
        \\
        LDMVFI~\cite{LDMVFI}
        & \cmark
        & 0.0145 & 0.0130 
        & 0.0284 & 0.0219
        & 0.0602 & 0.0379 
        & 0.1226 & 0.0651
        \\
        PerVFI~\cite{pervfi}
        & \cmark
        & 0.0142 & 0.0124
        & 0.0245 & 0.0181
        & 0.0561 & 0.0635
        & 0.0902 & 0.0448
        \\
        MoMo (Ours)
        & \cmark
        & \textbf{0.0111} & \textbf{0.0102}
        & \textbf{0.0202} & \textbf{0.0155}
        & \textbf{0.0419} & \textbf{0.0252}
        & \textbf{0.0872} & \textbf{0.0433}
        \\
    \bottomrule
    \end{tabular}
    \vspace{-0.5em}
    \caption{Quantitative experiments on the SNU-FILM benchmark~\cite{cain}.
    The best results are in \textbf{bold}, and the second best is \underline{underlined}, respectively. Our method outperforms existing methods on all four subsets.}
    \vspace{-0.5em}
    \label{tab:snufilm}
\end{table*}
\begin{table*}[t!]
    \centering
    \setlength{\tabcolsep}{1mm}
    \begin{tabular}{l c cc cc cc cc}
    \toprule
    
    \multirowcell{2}{Method} & \multirowcell{2}{Perception-\\oriented loss} & \multicolumn{2}{c}{Middlebury} & \multicolumn{2}{c}{Vimeo90k} & \multicolumn{2}{c}{Xiph-2K} & \multicolumn{2}{c}{Xiph-4K} \\
    && LPIPS & DISTS
    & LPIPS & DISTS
    & LPIPS & DISTS
    & LPIPS & DISTS \\
    
    \midrule\midrule
    
    ABME~\cite{abme}
    & \xmark
    & 0.0290 & 0.0325
    & 0.0213 & 0.0353
    & 0.1071 & 0.0581
    & 0.2361 & 0.1108
    \\
    XVFI$_v$~\cite{xvfi}
    & \xmark
    & 0.0169 & 0.0244
    & 0.0229 & 0.0354
    & 0.0844 & 0.0418
    & 0.1835 & 0.0779
    \\
    IFRNet-Large~\cite{ifrnet}
    & \xmark
    & 0.0285 & 0.0366
    & 0.0189 & 0.0325
    & 0.0681 & 0.0372
    & 0.1364 & 0.0665
    \\
    RIFE~\cite{rife}
    & \xmark
    & 0.0162 & 0.0228
    & 0.0223 & 0.0356
    & 0.0918 & 0.0481
    & 0.2072 & 0.0915
    \\
    FILM-$\mathcal{L}_1$~\cite{film}
    & \xmark
    & 0.0173 & 0.0246
    & 0.0197 & 0.0343
    & 0.0906 & 0.0510
    & 0.1841 & 0.0884
    \\
    AMT-G~\cite{amt}
    & \xmark
    & 0.0486 & 0.0533
    & 0.0195 & 0.0351
    & 0.1061 & 0.0563
    & 0.2054 & 0.1005
    \\
    EMA-VFI~\cite{emavfi}
    & \xmark
    & 0.0151 & 0.0218
    & 0.0196 & 0.0343
    & 0.1024 & 0.0550
    & 0.2258 & 0.1049
    \\
    UPRNet-LARGE~\cite{uprnet}
    & \xmark
    & 0.0150 & 0.0209
    & 0.0201 & 0.0342
    & 0.1010 & 0.0553
    & 0.2150 & 0.1017
    \\
    CAIN~\cite{cain}
    & \cmark
    & 0.0254 & 0.0383
    & 0.0306 & 0.0483
    & 0.1025 & 0.0533
    & 0.2229 & 0.0980
    \\
    FILM-${\mathcal{L}_{vgg}}$~\cite{film}
    & \cmark
    & 0.0096 & 0.0148
    & 0.0137 & 0.0229
    & 0.0355 & 0.0238
    & 0.0754 & 0.0406
    \\
    FILM-$\mathcal{L}_s$~\cite{film}
    & \cmark
    & \textbf{0.0093} & \underline{0.0140}
    & \textbf{0.0131} & \underline{0.0224}
    & \underline{0.0330} & 0.0237
    & \underline{0.0703} & 0.0385
    \\
    LDMVFI~\cite{LDMVFI}
    & \cmark
    & 0.0195 & 0.0261
    & 0.0233 & 0.0327
    & 0.0420 & 0.0163
    & 0.0859 & 0.0359
    \\
    PerVFI~\cite{pervfi}
    & \cmark
    & 0.0142 & 0.0163
    & 0.0179 & 0.0248
    & 0.0381 & \underline{0.0153}
    & 0.0858 & \underline{0.0331}
    \\
    MoMo (Ours)
    & \cmark
    & \underline{0.0094} & \textbf{0.0126}
    & \underline{0.0136} & \textbf{0.0203}
    & \textbf{0.0300} & \textbf{0.0119}
    & \textbf{0.0631} & \textbf{0.0274}
    \\
    
    \bottomrule
    \end{tabular}
    \vspace{-0.5em}
    \caption{Quantitative experiments on the three benchmarks, Middlebury~\cite{middlebury}, Vimeo90k~\cite{toflow} and Xiph-2K,4K~\cite{xiph,softsplat}.
    The best results are in \textbf{bold}, and the second best is \underline{underlined}, respectively.
    }
    \vspace{-0.5em}
    \label{tab:mb_vm_xiph}
\end{table*}

\subsection{Comparison to State-of-the-arts}
\paragraph{Baselines}
We compare our method, MoMo, with state-of-the-art VFI methods which employ perception-oriented objectives in the training process: CAIN~\cite{cain}, FILM-$\mathcal{L}_{vgg}$, FILM-$\mathcal{L}_s$~\cite{film}, LDMVFI~\cite{LDMVFI} and PerVFI~\cite{pervfi}.
Since there is a limited number of methods focused on perception-oriented objectives,
we also include methods trained with the traditional pixel-wise reconstruction loss for comparison: XVFI~\cite{xvfi}, RIFE~\cite{rife}, IFRNet~\cite{ifrnet}, AMT~\cite{amt}, EMA-VFI~\cite{emavfi}, UPRNet~\cite{uprnet}.

\subsubsection{Quantitative Results}
Tables~\ref{tab:snufilm} and \ref{tab:mb_vm_xiph} present our quantitative results across four benchmark datasets.
MoMo achieves state-of-the-art on all four subsets of SNU-FILM, leading in both LPIPS and DISTS metrics. On Middlebury and Vimeo90k, it outperforms baselines in DISTS and closely trails FILM-$\mathcal{L}_s$ in LPIPS. MoMo also excels on both Xiph subsets, 2K and 4K, in both metrics.
This highlights the effectiveness of our approach in generating well-structured optical flows for the intermediate frame through proficient intermediate motion modeling.

\begin{figure}[t!]
    \centering
    \includegraphics[width=\linewidth]{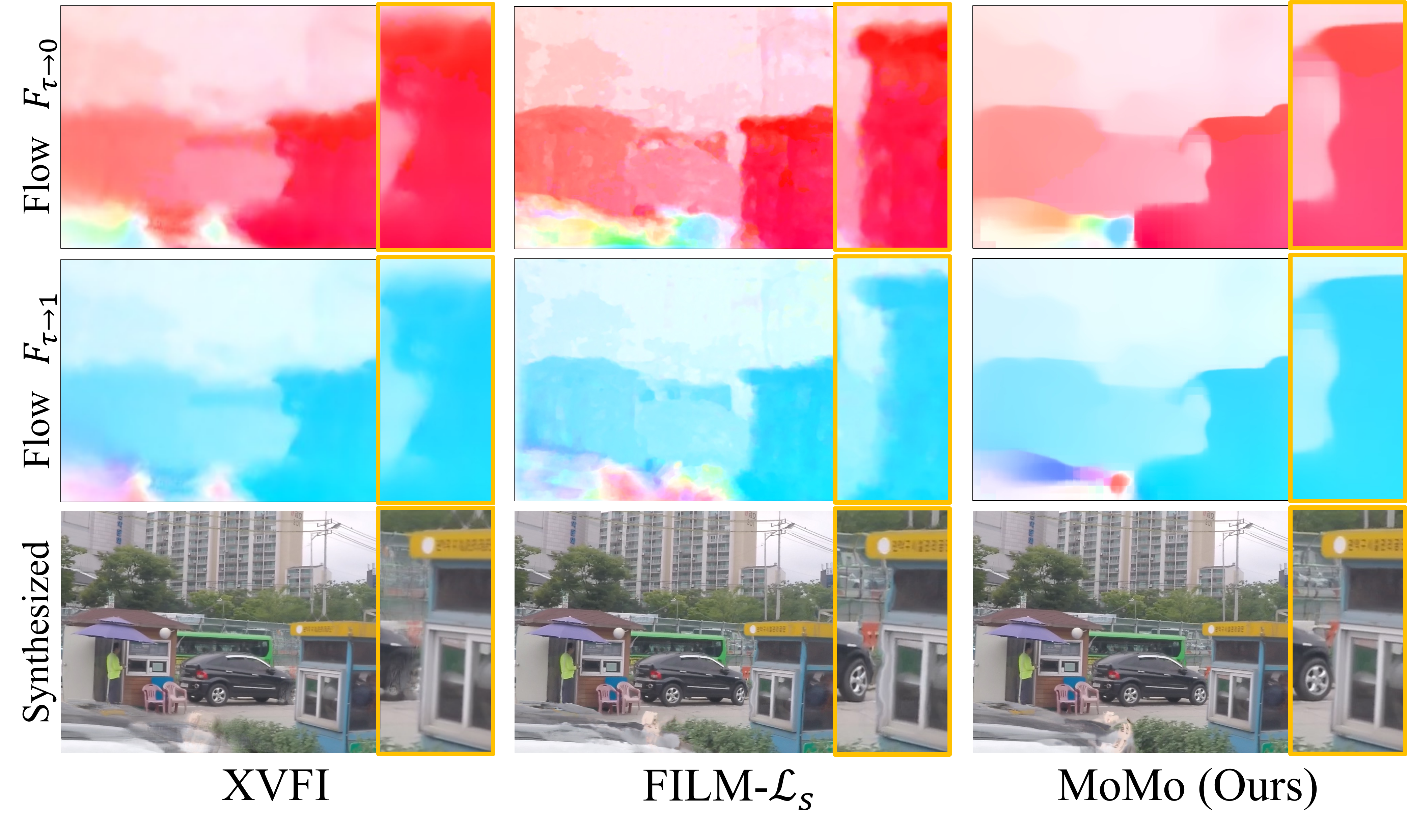}
    \vspace{-2em}
    \caption{Visualized comparison of estimated intermediate flows against state-of-the-art methods. Our flow estimations show better-structured flow fields which leads to promising synthesis of frames.
    }
    \vspace{-1em}
    \label{fig:flow_comparison}
\end{figure}

To support our discussion, we provide a visualization of flow estimations and the frame synthesis outcomes, with comparison to state-of-the-art algorithms at Fig.~\ref{fig:flow_comparison}.
Although XVFI and FILM takes advantage of the recurrent architecture tailored for flow estimations at high resolution images, they fail in well-structured flow estimations and frame synthesis. XVFI largely fails in flow estimation, which results in blurry outputs.
The estimations by FILM display vague and noisy motion boundaries, especially in $F_{\tau\rightarrow1}$. Another important point to note is that the flow pair $F_{\tau\rightarrow0}$ and $F_{\tau\rightarrow1}$ of FILM do not align well with each other, causing confusion in the synthesis process.

\begin{figure}[h!]
    \centering
    \includegraphics[width=\linewidth]{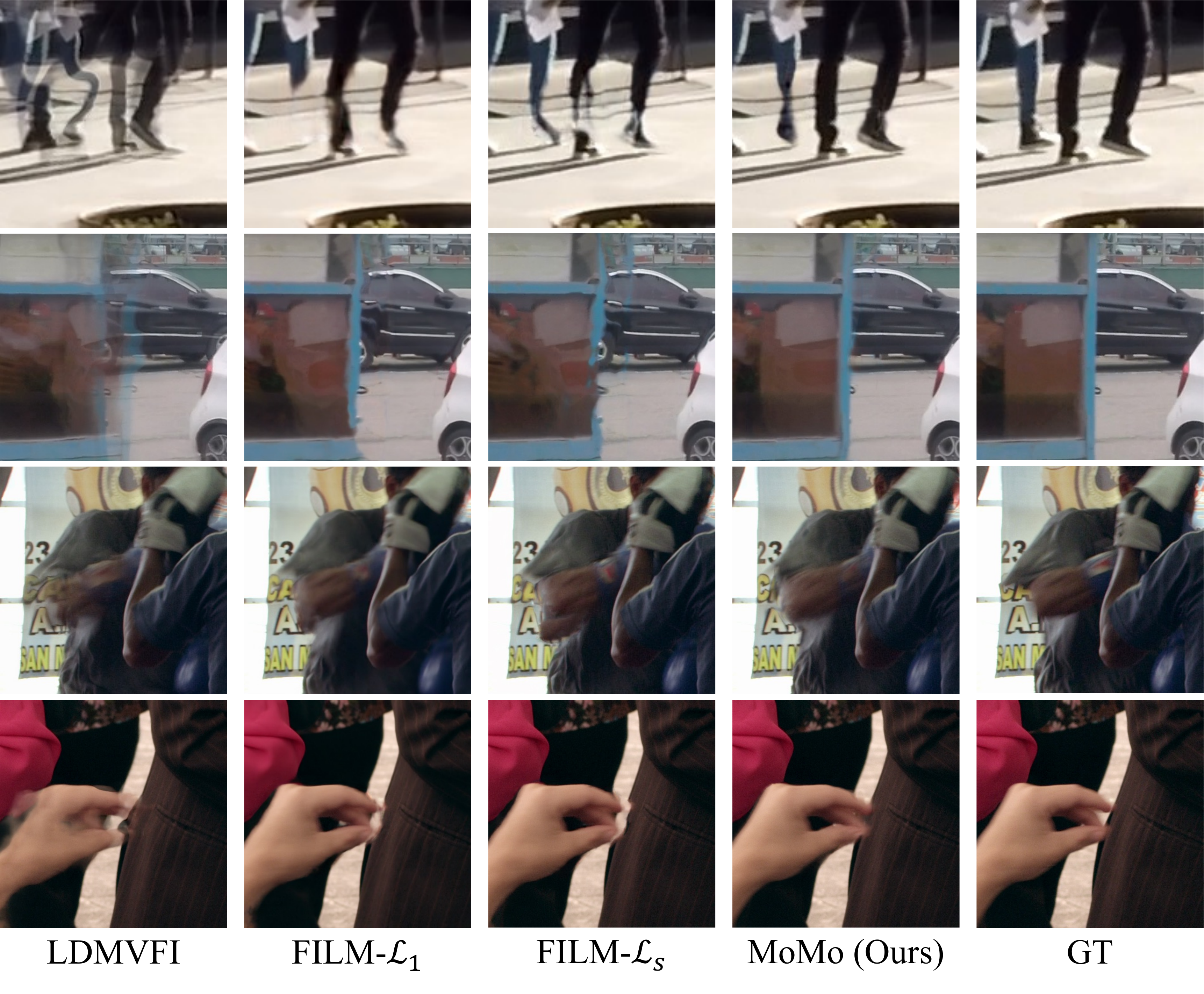}
    \vspace{-2em}
    \caption{Qualitative comparison against state-of-the-art methods on `extreme' subset of SNU-FILM and Xiph-4K. Our results show the least artifacts and generate well-structured images.}
    \vspace{-1em}
    \label{fig:qualitative}
\end{figure}

\subsubsection{Qualitative Results}
The qualitative results of MoMo with comparison to the state-of-the-art algorithms can be found at Fig.~\ref{fig:main} and \ref{fig:qualitative}.
In Fig.~\ref{fig:main}, MoMo reconstructs both wings with rich details, whereas methods of the top row, which greatly relies on the $L_1$ pixel-wise loss in training show blurry results.
Moreover, our result also outperforms state-of-the-art models designed particularly for perceptual quality, LDMVFI and FILM-$\mathcal{L}_s$, with well-structurized synthesis of both wings.
Fig.~\ref{fig:qualitative} present additional results obtained from the `extreme' subset of SNU-FILM and Xiph-4K set.
MoMo consistently shows a superior visual quality, with less artifacts and well-structured objects.
The interpolated video samples are provided in the supplementary material.

\subsection{Ablation Studies}
\label{sec:ablation}

We conduct ablation studies to verify the effects of our design choices. We use the `hard' subset of SNU-FILM dataset, unless mentioned otherwise.
We start by studying the effects of the teacher optical flow model, used for training the motion diffusion model.
We then experiment on the number of denoising steps used at inference time.
Lastly, we study on the design choices in diffusion architecture.

\begin{table*}[t!]
    \centering
    \setlength{\tabcolsep}{1mm}
    \begin{tabular}{c l l cc c c c}
    \toprule
        Data& Prediction & Architecture & LPIPS & DISTS & TFLOPs & Params. (M) & R-time (ms)\\
        \midrule\midrule
        Latent & $\epsilon$ & Standard U-Net~\cite{LDMVFI} & 0.0601 & 0.0379 & 3.25 & 439.0 & 10283.51\\
        \midrule
         Flow & $\epsilon$ & Standard U-Net & 0.4090 & 0.2621 & 8.08 & 71.1 & 603.64\\
         Flow & $\textbf{x}_0$ & Standard U-Net & 0.0460 & 0.0295 & 8.08 & 71.1 & 603.64\\
         Flow & $\textbf{x}_0$ (weighted) & 
         Convex-Up U-Net (Ours) & 0.0463 & 0.0298 & \textbf{1.12} & \textbf{73.6} & \textbf{145.49}\\
         Flow & $\textbf{x}_0$ &
         Convex-Up U-Net (Ours) & \textbf{0.0425} & \textbf{0.0257} & \textbf{1.12} & \textbf{73.6} & \textbf{145.49}\\
         \midrule
         Flow & $\textbf{x}_0$ &
         Convex-Up U-Net (Ours) + Longer Training & 0.0419 & 0.0252 & 1.12 & 73.6 & 145.49\\
    \bottomrule
    \end{tabular}
    \vspace{-0.8em}
    \caption{
    Ablation study on our motion diffusion model. 
    Our design choice reaches the best performance with minimal computational needs and fastest runtime.
    The first row includes our baseline, LDMVFI~\cite{LDMVFI}, for reference.
    }
    \vspace{-1.2em}
    \label{tab:ablation_arch}
\end{table*}

\subsubsection{Optical Flow Teacher}
We conduct a study on the teacher optical flow model.
We choose RAFT~\cite{raft} as the teacher model, the state-of-the-art model for optical flow estimation.
We test the teacher model of three different weights:
1) the default off-the-shelf weights provided by the Torchvision library~\cite{torchvision}.
2) Initialized with the pre-trained weights, weights trained jointly with our synthesis network in an end-to-end manner.
3) Optimized in an alternating manner with the synthesis network, initialized from pre-trained weights, as described in Sec.~\ref{sec:synth_and_teacher}.
We use these three different versions of RAFT as the teacher for the experiment.

The ablation study summarized in Table~\ref{tab:ablation_teacher} shows that fine-tuning the flow model $\mathcal{F}$ after training the synthesis model $\mathcal{G}$ is the most effective.
Fine-tuning of $\mathcal{F}$ enhances flow estimation and suitability for synthesis tasks~\cite{toflow}.
However, end-to-end training can cause the synthesis model to depend too heavily on estimated flows, risking inaccuracies from the motion diffusion model. Our results highlight that sequential training of the synthesis model and the flow estimator ensures optimal performance.

\subsubsection{Diffusion Architecture}
In our ablation study, detailed in Table~\ref{tab:ablation_arch}, we assess our motion diffusion model using the standard timestep-conditioned U-Net architecture (\texttt{UNet2DModel}) from the diffusers library~\cite{diffusers}, alongside $\epsilon$- and $\textbf{x}_0$-prediction types. Contrary to the common preference for $\epsilon$-prediction in diffusion models, our motion diffusion model favors $\textbf{x}_0$-prediction.

We also explore our coarse-to-fine estimation using convex upsampling. This approach reduces computational costs and improves performance. Given that our architecture predicts values with a strong correlation between neighboring pixels, $\epsilon$-prediction, which samples noise independently, proves less suitable.
We experiment with a SNR-weighted $\textbf{x}_0$-prediction~\cite{progressivedistill}, to make it equivalent to $\epsilon$-prediction loss. Nonetheless, $\textbf{x}_0$-prediction consistently outperforms, validating our architectural decisions.

Despite having a similar number of parameters as the standard U-Net, our Convex Upsampling U-Net significantly reduces floating point operations (FLOPs) by about $7.2\times$. Runtime tests on a NVIDIA 32GB V100 GPU for $256\times448$ resolution frames—averaged over 100 iterations—reveal that our Convex-Up U-Net processes frames in approximately 145.49 ms each, achieving a $4.15\times$ speedup over the standard U-Net and an $70\times$ faster inference speed than the LDMVFI baseline. This efficiency is attributed to our model's efficient architecture and notably fewer denoising steps.

\begin{table}[t!]
    \centering
    \setlength{\tabcolsep}{1mm}
    \begin{tabular}{l  cc cc}
        \toprule
        Teacher & Fine-tune $\mathcal{F}$ & Train $\mathcal{S}$& LPIPS & DISTS \\
        \midrule\midrule
         Pre-trained & \xmark & \xmark & 0.0445 & 0.0284 \\
         End-to-End & \cmark & \cmark & 0.0475 & 0.0287 \\
         Alternating (Ours) & \cmark & \xmark & \textbf{0.0419} & \textbf{0.0252} \\
    \bottomrule
    \end{tabular}
    \vspace{-0.5em}
    \caption{Experiments on the teacher flow model. We use RAFT~\cite{raft} with three different weights.
    The results show that alternating optimization, fine-tuning the flow model with $\mathcal{S}$ fixed, to be the most effective.}
    \label{tab:ablation_teacher}
\end{table}

\begin{table}[t!]
    \centering
    \begin{tabular}{l cc}
    \toprule
        \# of steps & LPIPS & DISTS\\
        \midrule\midrule
        1 step ($\approx$ non-diffusion)  & 0.0892 & 0.0452 \\
        8 step (default)        & \textbf{0.0872} & \textbf{0.0433} \\
        20 step                 & 0.0872 & 0.0433 \\
        50 step                 & 0.0874 & 0.0435 \\
    \bottomrule
    \end{tabular}
    \vspace{-0.5em}
    \caption{Experiment on the number of denoising steps for inference (on SNU-FILM-extreme). Our experiments show that about 8 steps is enough, and use of more steps exceeding this does not lead to a notable improvement.}
    \vspace{-1em}
    \label{tab:ablation_steps}
\end{table}

\subsubsection{Effectiveness of Diffusion}
Table~\ref{tab:ablation_steps} shows the effect of number of denoising steps in motion generation, experimented on the `extreme' subset of SNU-FILM.
We observe consistent improvement with more number of steps up to 8, with more steps not markedly improving performance.
In contrast to image diffusion models~\cite{ldm} and LDMVFI~\cite{LDMVFI} which requires over 50 steps,
our method delivers satisfactory outcomes with far fewer steps, cutting down on both runtime and computational expenses.
This is likely due to the simpler nature of flow representations compared to RGB pixels.
This experiment shows the effectiveness of using diffusion models in motion modeling, as use of multiple steps guarantees better motion predictions.

\section{Conclusion}
In this paper, we proposed MoMo, a disentangled motion modeling framework for perceptual video frame interpolation.
Our approach mainly focuses on modeling the intermediate motions between frames, with explicit supervision on the motions only.
We introduced motion diffusion model, which generates intermediate bi-directional flows necessary to synthesize the target frame with a novel architecture tailored for optical flow generation, which greatly improves both performance and computational efficiency.
Extensive experiments confirm that our method achieve state-of-the-art quality on multiple benchmarks.

\section*{Acknowledgement}
This work was partly supported by Institute of Information \& communications Technology Planning \& Evaluation (IITP) grant funded by the Korea government(MSIT) [NO.RS-2021-II211343, Artificial Intelligence Graduate School Program (Seoul National University)],
the National Research Foundation of Korea (NRF) grant
funded by the Korea government (MSIT) (No.  2022R1A3B1077720),
the National Research Foundation of
Korea (NRF) grant funded by the Korea government (MSIT)(No. 2022R1A5A7083908),
and the BK21 FOUR program of the Education and the Research Program for Future ICT Pioneers, Seoul National University in 2024.

\bibliography{aaai25}
\clearpage
\appendix
\setcounter{table}{0}
\renewcommand{\thetable}{A\arabic{table}}
\setcounter{figure}{0}
\renewcommand{\thefigure}{A\arabic{figure}}

\section{Implementation Details}
\subsection{Recurrent Synthesis}
We build our synthesis network $\mathcal{S}$ to be of recurrent structure, motivated by the recent trend in video frame interpolation~\cite{xvfi, uprnet, film}, due to its great efficiency.
Let the number of recurrent process be $L-1$, and $\mathcal{S}$ can be expressed as recurrent application of a synthesis process $\mathcal{P}$:
\begin{equation}
    \mathcal{S}(I_{in}, F_\tau) = \mathcal{P}^0(\cdots \mathcal{P}^{L-1}(\hat{I}^{L}_\tau, I^{L-1}_{in}, F^{L-1}_\tau)\cdots, I^0_{in}, F^0_\tau),
\end{equation}
where $I_{in}^l$ denotes the input image pair $I_0, I_1$ downsampled by a factor of $2^l\times$, and $F_\tau^l$ denotes the flow map pair downsampled likewise.

Our process $\mathcal{P}^l$ at level $l$ is described as follows.
$\mathcal{P}^l$ takes three components as the input: 1) frame $\hat{I}^{l+1}_\tau$, synthesized from the previous level $l+1$, 2) downsampled input frame pair $I^l_{in}=\{I_0^l, I_1^l\}$, and 3) the downsampled flow maps $F_\tau=\{ F_{\tau\rightarrow0}, F_{\tau\rightarrow1}\}$.
First, using the input frame pair $I^l_{in}=\{I_0^l, I_1^l\}$ and its corresponding flow pair $F_\tau=\{ F_{\tau\rightarrow0}, F_{\tau\rightarrow1}\}$,
we perform backward-warping $(\overleftarrow{\omega})$ on the two frames with their corresponding flows:
\begin{equation}
    I^l_{\tau \leftarrow i} = \overleftarrow{\omega}(I^l_i, F^l_{\tau \rightarrow i}),
    i \in \{0, 1\}.
\end{equation}

Next, we take frame $\hat{I}^{l+1}_\tau$ and use bicubic upsampling to match the size of level $l$, denoted as $\hat{I}^{l+1\rightarrow l}_\tau$.
The upsampled frame $\hat{I}^{l+1\rightarrow l}_\tau$, along with the two warped frames $I^l_{\tau \leftarrow 0}, I^l_{\tau \leftarrow 1}$ and their corresponding flow maps $F^l_{\tau \rightarrow 0}, F^l_{\tau \rightarrow 1}$ are given to the synthesis module $\mathcal{G}$, which outputs a 4 channel output --- 1 channel occlusion mask $M_0$ to blend $I_{\tau\leftarrow 0}$ and $I_{\tau\leftarrow 1}$, and 3 channel residual RGB values $\Delta\hat{I_\tau}$:

\begin{equation}
    M^l_0, \Delta\hat{I}^l_\tau = \mathcal{G}(\hat{I}^{l+1\rightarrow l}_\tau, I^l_{\tau\leftarrow0}, I^l_{\tau\leftarrow1}, F^l_{\tau \rightarrow 0}, F^l_{\tau \rightarrow 1}).
\end{equation}

Using these outputs, we obtain the output of $\mathcal{P}^l$, the synthesized frame at level $l$:
\begin{equation}
    \mathcal{P}^l(\hat{I}^{l+1}_\tau, I_{in}^l, F_\tau^l) = I^l_{\tau\leftarrow 0} \odot M^l_0 + I^l_{\tau\leftarrow 1} \odot (1 - M^l_0) + \Delta\hat{I}^l_\tau .
\end{equation}

Note that $\hat{I}^L_\tau$ is not available for level $\mathcal{P}^{L-1}$, since it is of the highest level. Therefore we equally blend the two warped frames at level $l=L-1$ as a starting point: $\hat{I}^{L\rightarrow L-1}_\tau = I^{L-1}_{\tau\leftarrow0} \odot 0.5 + I^{L-1}_{\tau\leftarrow1} \odot 0.5$.

\subsection{Perception-oriented Loss}
As mentioned in Sec.
\ref{sec:objective},
we specify the objective function we use for stage 1 training.
Along with the $L_1$ pixel reconstruction loss, we use an LPIPS-based perceptual loss,
which computes the $L_2$ distance in the deep feature space of AlexNet~\cite{alexnet}.
This loss is well-known for high correlation with human judgements.
Next, we exploit the style loss~\cite{gatysstyle} $\mathcal{L}_G$ as its effectiveness has been proved in a recent work~\cite{film}. This loss computes the $L_2$ distance of feature correlations extracted from the VGG-19 network~\cite{vgg}:
\begin{equation}
    \mathcal{L}_G=\frac{1}{N} \sum^N_{n=1} \alpha_n ||G_n(I_\tau) - G_n(\hat{I}_\tau)||_2.
\end{equation}
Here, $\alpha_n$ denotes the weighting hyper-parameter of the $n$-th selected layer.
Denoting the feature map of frame $I_\tau$ extracted from $n$-th selected layer of the VGG~\cite{vgg} network as $\phi_n(I_\tau) \in \mathbb{R}^{H\times W \times C}$,
the Gram matrix of frame $I_\tau$ at the $n$-th feature space, $G_n(I_\tau)\in \mathbb{R}^{C\times C}$ can be acquired as follows:
\begin{equation}
    G_n(I_\tau) = \phi_n(I_\tau)^\top \phi_n(I_\tau).
    \label{eq:gram}
\end{equation}
Likewise, the Gram matrix of our synthesized frame, $G_n(\hat{I}_\tau)$, could be computed by substituting $I_\tau$ with $\hat{I}_\tau$ in Eq.~\ref{eq:gram}.

\subsection{Training Details}
We elaborate on our training details mentioned in Sec.~\ref{sec:impl_details}.

\paragraph{Stage 1 Training}
We employ the AdamW optimizer~\cite{adamw}, setting the weight decay to $10^{-4}$ and the batch size to 32 both in the training process of $\mathcal{G}$ and $\mathcal{F}$.
We train the synthesis model $\mathcal{G}$ for a total of 200 epochs, with a fixed learning rate of $2\times10^{-4}$.
For the first 150 epochs, we set the hyper-parameters to $\lambda_1=1, \lambda_p=0, \lambda_G=0$. After that, we use $\lambda_1=1, \lambda_p=1, \lambda_G=20$ for the last 50 epochs.
Once the synthesis model is fully trained, we fine-tune the teacher flow model $\mathcal{F}$ for 100 epochs, with its learning rate fixed to $10^{-4}$. We set hyper-parameters to $\lambda_1=1, \lambda_p=1, \lambda_G=20$.
Both $\mathcal{G}$ and $\mathcal{F}$ benefit from an exponential moving average (EMA) with a 0.999 decay rate.
We set the number of pyramids $L=3$ during training, and use $L=\lceil \log_{2}(R/32) \rceil$ for resolution $R$ at inference.

\paragraph{Stage 2 Training}
We train our diffusion model for 500 epochs using the AdamW optimizer with a constant learning rate of $2\times10^{-4}$, weight decay of $10^{-8}$, and batch size of 64, applying an EMA with a 0.9999 decay rate. Given that diffusion models typically operate with data values between $[-1, 1]$ but optical flows often exceed this range, we normalize flow values by dividing them by 128. This adjustment ensures flow values to be compatible with the diffusion model's expected data range, effectively aligning flow values with those of the RGB space, which are similarly normalized. We utilize a linear noise schedule~\cite{ddpm} and perform 8 denoising steps using the ancestral DDPM sampler~\cite{ddpm} for efficient sampling.

\subsection{Inference}
Since our motion diffusion model is trained on a well-curated data ranging within 256 resolution, it could suffer from a performance drop when it comes to high resolution videos of large motions which goes beyond the distribution of the training data.
To handle these cases, we generate flows at the training resolution by resizing the inputs, followed by post-processing of bicubic upsampling at inference time.

\subsection{Input Downsampling}
\begin{figure}
    \centering
    \includegraphics[width=0.75\linewidth]{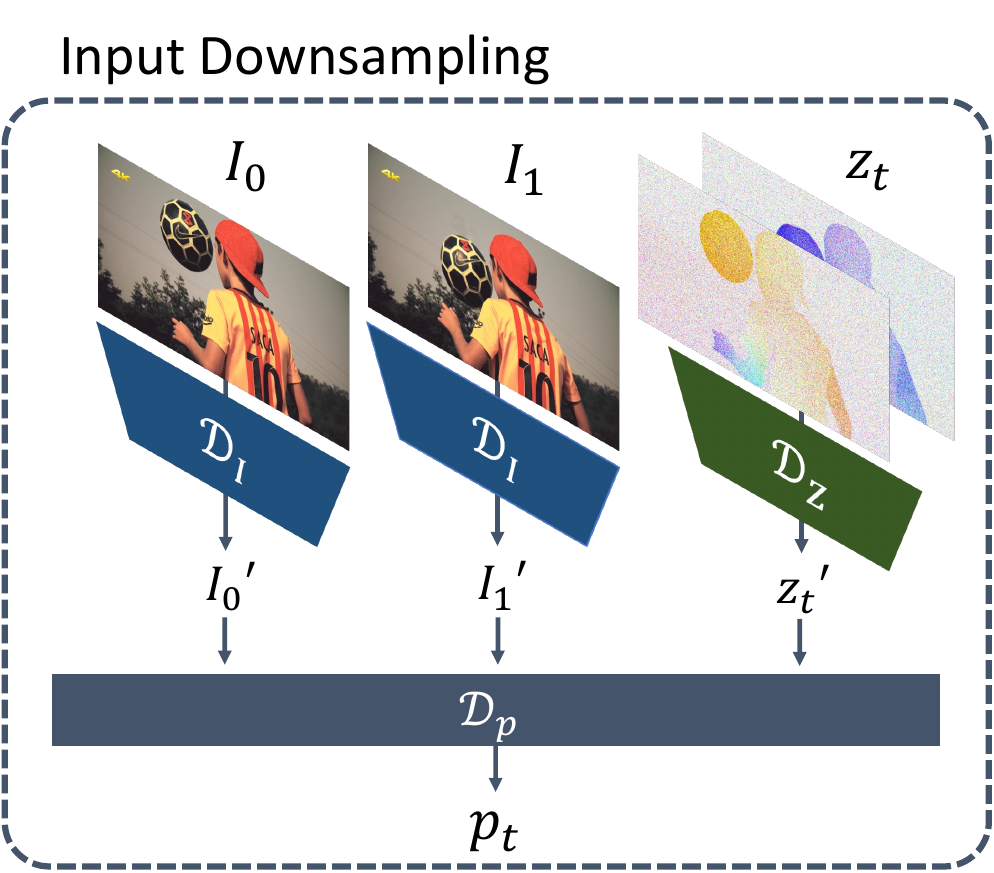}
    \caption{Visualized description of input downsampling in our motion diffusion model.}
    \label{fig:input_down}
\end{figure}

We provide illustrated description of input downsampling (Sec.~\ref{sec:motion_modeling}) in Fig.~\ref{fig:input_down}.

\subsection{Convex Upsampling}
\begin{figure}[t!]
    \centering
    \vspace{-8pt}
    \includegraphics[width=\linewidth]{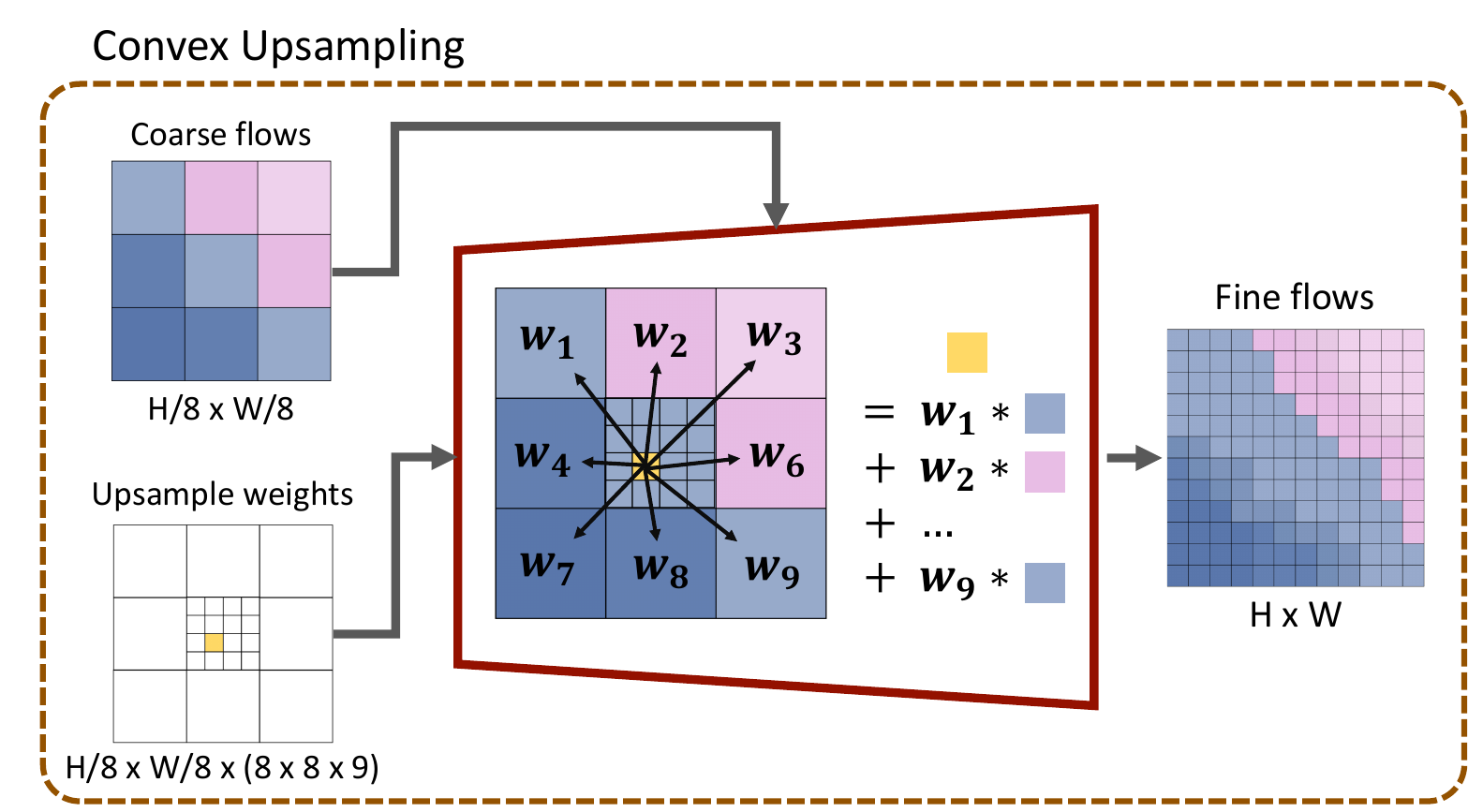}
    \caption{Visualization of convex upsampling layer. The dimensions of flows are omitted for simplicity. The upsampling weights are predicted for both directions and applied to the bi-directional flows in the same manner.
    }
    \vspace{-12pt}
    \label{fig:convex_up}
\end{figure}

We provide a illustrated description of convex upsampling (Sec.~\ref{sec:motion_modeling}) in Fig.~\ref{fig:convex_up}.

\section{Additional Experiments}
\subsection{Further Analysis on Denoising Steps}
We experiment on the effect of different number of denoising steps for motion modeling, on the `hard' subset of SNU-FILM.(Tab.\ref{tab:ablation_steps_full}) Although the performance does improve up to 8 steps, the increase is relatively marginal compared to the results on the `extreme' subset.
We speculate the reason for this result is due to the smaller ill-posedness of the `hard' subset, which limits the diversity of feasible flows.
We claim that the use of more steps and the design choice of diffusion models for motion modeling is more advantageous as the ill-posedness of motions gets larger.

\begin{table}[t!]
    \centering
    \resizebox{\linewidth}{!}{
    \begin{tabular}{l cccc}
    \toprule
        \multirowcell{2}{\# of steps} & \multicolumn{2}{c}{SNU-FILM-hard} & \multicolumn{2}{c}{SNU-FILM-extreme}\\
         & LPIPS & DISTS & LPIPS & DISTS\\
        \midrule\midrule
        1 step              & 0.0421 & 0.0254 & 0.0892 & 0.0452 \\
        8 step (default)    & \textbf{0.0419} & \textbf{0.0252} & \textbf{0.0872} & \textbf{0.0433} \\
        20 step             & 0.0420 & 0.0253 & 0.0872 & 0.0433 \\
        50 step             & 0.0420 & 0.0254 & 0.0874 & 0.0435 \\
    \bottomrule
    \end{tabular}
    }
    \caption{Experiment on the number of denoising steps at inference time. Our experiments show that about 8 steps is enough, and use of more steps exceeding this does not lead to a notable improvement considering the runtime tradeoff.}
    \label{tab:ablation_steps_full}
\end{table}

\subsection{Full Quantitative Results}
We report the full quantitative results including the fidelity metrics such as PSNR and SSIM on the SNU-FILM (Tab.~\ref{tab:snufilm_full_easy_medium},~\ref{tab:snufilm_full_hard_extreme}), Middlebury, Vimeo90k (Tab.~\ref{tab:mb_vm_full}) and Xiph benchmarks (Tab.~\ref{tab:xiph_full}).

In addition to the fidelity metrics, we also include the results of a lighter version of our model with 10M parameters, denoted as \textbf{MoMo-10M}.

\begin{table*}[t!]
    \centering
    \setlength{\tabcolsep}{1mm}
    
    \begin{tabular}{l c cccc cccc}
    
    \toprule
    
        \multirowcell{2}{Method} & \multirowcell{2}{Perception-\\oriented loss} & \multicolumn{4}{c}{SNU-FILM-easy} & \multicolumn{4}{c}{SNU-FILM-medium}\\
        && PSNR & SSIM & LPIPS & DISTS
        & PSNR & SSIM & LPIPS & DISTS\\
        
    \midrule\midrule
        
        ABME~\cite{abme} 
        & \xmark
        & 39.59 & 0.9901 & 0.0222 & 0.0229 
        & 35.77 & 0.9789 & 0.0372 & 0.0344 
        \\
        XVFI$_v$~\cite{xvfi}
        & \xmark
        & 39.78 & 0.9865 & 0.0175 & 0.0181
        & 35.36 & 0.9692 & 0.0322 & 0.0276
        \\
        IFRNet-Large~\cite{ifrnet} 
        & \xmark
        & \underline{40.10} & 0.9906 & 0.0203 & 0.0211
        & \underline{36.12} & \underline{0.9797} & 0.0321 & 0.0288
        \\
        RIFE~\cite{rife} 
        & \xmark
        & 40.06 & \underline{0.9907} & 0.0181 & 0.0195
        & 35.75 & 0.9789 & 0.0317 & 0.0289
        \\
        FILM-$\mathcal{L}_1$~\cite{film} 
        & \xmark
        & 39.74 & 0.9902 & 0.0184 & 0.0217
        & 35.81 & 0.9789 & 0.0315 & 0.0316
        \\
        AMT-G~\cite{amt} 
        & \xmark
        & 38.47 & 0.9880 & 0.0325 & 0.0312
        & 35.39 & 0.9779 & 0.0447 & 0.0395
        \\
        EMA-VFI~\cite{emavfi} 
        & \xmark
        & 39.52 & 0.9903 & 0.0186 & 0.0204
        & 35.83 & 0.9795 & 0.0325 & 0.0318
        \\ 
        UPRNet-LARGE~\cite{uprnet} 
        & \xmark
        & \textbf{40.44} & \textbf{0.9911} & 0.0182 & 0.0203
        & \textbf{36.29} & \textbf{0.9801} & 0.0334 & 0.0327
        \\
        CAIN~\cite{cain}
        & \cmark
        & 39.89 & 0.9900 & 0.0197 & 0.0229  
        & 35.61 & 0.9776 & 0.0375 & 0.0347
        \\
        FILM-$\mathcal{L}_{vgg}$~\cite{film}
        & \cmark
        & 39.79 & 0.9900 & 0.0123 & 0.0128
        & 35.77 & 0.9782 & 0.0219 & 0.0183
        \\
        FILM-$\mathcal{L}_s$~\cite{film}
        & \cmark
        & 39.68 & 0.9900 & \underline{0.0120} & \underline{0.0124}
        & 35.70 & 0.9781 & \underline{0.0213} & \underline{0.0177}
        \\
        LDMVFI~\cite{LDMVFI}
        & \cmark
        & 38.68 & 0.9834 & 0.0145 & 0.0130 
        & 33.90 & 0.9703 & 0.0284 & 0.0219
        \\
        PerVFI~\cite{pervfi}
        & \cmark
        & 38.02 & 0.9831 & 0.0142 & \underline{0.0124}
        & 34.57 & 0.9662 & 0.0245 & 0.0181
        \\
        
        MoMo (Ours)
        & \cmark
        & 39.64 & 0.9895 & \textbf{0.0111} & \textbf{0.0102}
        & 35.45 & 0.9769 & \textbf{0.0202} & \textbf{0.0155}
        \\
        MoMo-10M (Ours)
        & \cmark
        & 39.54 & 0.9896 & 0.0111 & 0.0103
        & 35.36 & 0.9769 & 0.0204 & 0.0157
        \\

    \bottomrule
    \end{tabular}
    \caption{Full quantitative results including the fidelity metrics (PSNR, SSIM) on the `easy' and `medium' subsets of SNU-FILM benchmark~\cite{cain}
    The best results are in \textbf{bold}, and the second best is \underline{underlined}, respectively.}
    \label{tab:snufilm_full_easy_medium}
\end{table*}

\begin{table*}[t!]
    \centering
    \setlength{\tabcolsep}{1mm}
    
    \begin{tabular}{l c cccc cccc}
    
    \toprule
    
        \multirowcell{2}{Method} & \multirowcell{2}{Perception-\\oriented loss} & \multicolumn{4}{c}{SNU-FILM-hard} & \multicolumn{4}{c}{SNU-FILM-extreme}\\
        && PSNR & SSIM & LPIPS & DISTS
        & PSNR & SSIM & LPIPS & DISTS\\
        
        \midrule\midrule
        
        ABME~\cite{abme} 
        & \xmark
        & 30.58 & 0.9364 & 0.0658 & 0.0496 
        & 25.42 & 0.8639 & 0.1258 & 0.0747
        \\
        XVFI$_v$~\cite{xvfi}
        & \xmark
        & 29.91 & 0.9073 & 0.0629 & 0.0414
        & 24.67 & 0.8092 & 0.1257 & 0.0673
        \\
        IFRNet-Large~\cite{ifrnet} 
        & \xmark
        & 30.63 & 0.9368 & 0.0562 & 0.0403
        & 25.27 & 0.8609 & 0.1131 & 0.0638
        \\
        RIFE~\cite{rife} 
        & \xmark
        & 30.10 & 0.9330 & 0.0657 & 0.0443
        & 24.84 & 0.8534 & 0.1390 & 0.0764
        \\
        FILM-$\mathcal{L}_1$~\cite{film} 
        & \xmark
        & 30.42 & 0.9353 & 0.0568 & 0.0441
        & 25.17 & 0.8593 & 0.1060 & 0.0632
        \\
        AMT-G~\cite{amt} 
        & \xmark
        & 30.70 & \underline{0.9381} & 0.0680 & 0.0506 
        & \textbf{25.64} & \textbf{0.8658} & 0.1128 & 0.0686
        \\
        EMA-VFI~\cite{emavfi} 
        & \xmark
        & \underline{30.79} & \textbf{0.9386} & 0.0579 & 0.0457
        & 25.59 & \underline{0.8648} & 0.1099 & 0.0671
        \\ 
        UPRNet-LARGE~\cite{uprnet} 
        & \xmark
        & \textbf{30.86} & 0.9377 & 0.0612 & 0.0475
        & \underline{25.63} & 0.8641 & 0.1109 & 0.0672
        \\
        CAIN~\cite{cain}
        & \cmark
        & 29.90 & 0.9292 & 0.0885 & 0.0606
        & 24.78 & 0.8507 & 0.1790 & 0.1042
        \\
        FILM-$\mathcal{L}_{vgg}$~\cite{film}
        & \cmark
        & 30.34 & 0.9332 & 0.0443 & 0.0282
        & 25.11 & 0.8557 & 0.0917 & 0.0471
        \\
        FILM-$\mathcal{L}_s$~\cite{film}
        & \cmark
        & 30.29 & 0.9329 & \underline{0.0429} & \underline{0.0268}
        & 25.07 & 0.8550 & \underline{0.0889} & \underline{0.0448}
        \\
        LDMVFI~\cite{LDMVFI}
        & \cmark
        & 28.51 & 0.9173 & 0.0602 & 0.0379 
        & 23.92 & 0.8372 & 0.1226 & 0.0651
        \\
        PerVFI~\cite{pervfi}
        & \cmark
        & 29.68 & 0.9287 & 0.0561 & 0.0635
        & 25.03 & 0.8120 & 0.0902 & \underline{0.0448}
        \\
        MoMo (Ours)
        & \cmark
        & 30.12 & 0.9312 & \textbf{0.0419} & \textbf{0.0252}
        & 25.02 & 0.8547 & \textbf{0.0872} & \textbf{0.0433}
        \\
        MoMo-10M (Ours)
        & \cmark
        & 30.00 & 0.9308 & 0.0425 & 0.0257
        & 24.91 & 0.8535 & 0.0882 & 0.0438
        \\

    \bottomrule
     
    \end{tabular}
    \caption{Full quantitative results including the fidelity metrics (PSNR, SSIM) on the `hard' and `extreme' subsets of SNU-FILM benchmark~\cite{cain}.
    The best results are in \textbf{bold}, and the second best is \underline{underlined}, respectively.}
    \label{tab:snufilm_full_hard_extreme}
\end{table*}
\begin{table*}[t!]
    \centering
    \setlength{\tabcolsep}{1mm}
    
    \begin{tabular}{l c cccc cccc}
    
    \toprule
    
    \multirowcell{2}{Method} & \multirowcell{2}{Perception-\\oriented loss} & \multicolumn{4}{c}{Middlebury} & \multicolumn{4}{c}{Vimeo90k}\\
    && PSNR & SSIM & LPIPS & DISTS
    & PSNR & SSIM & LPIPS & DISTS \\
    
    \midrule\midrule
    
    ABME~\cite{abme}
    & \xmark
    & 37.05 & 0.9845 & 0.0290 & 0.0325
    & 36.18 & 0.9805 & 0.0213 & 0.0353
    \\
    XVFI$_v$~\cite{xvfi}
    & \xmark
    & 36.72 & 0.9826 & 0.0169 & 0.0244
    & 35.07 & 0.9710 & 0.0229 & 0.0354
    \\
    IFRNet-Large~\cite{ifrnet}
    & \xmark
    & 36.27 & 0.9816 & 0.0285 & 0.0366
    & 36.20 & 0.9808 & 0.0189 & 0.0325
    \\
    RIFE~\cite{rife}
    & \xmark
    & 37.16 & 0.9853 & 0.0162 & 0.0228
    & 35.61 & 0.9780 & 0.0223 & 0.0356
    \\
    FILM-$\mathcal{L}_1$~\cite{film}
    & \xmark
    & 37.37 & 0.9838 & 0.0173 & 0.0246
    & 35.89 & 0.9796 & 0.0197 & 0.0343
    \\
    AMT-G~\cite{amt}  
    & \xmark
    & 34.23 & 0.9708 & 0.0486 & 0.0533
    & \textbf{36.53} & \textbf{0.9819} & 0.0195 & 0.0351
    \\
    EMA-VFI~\cite{emavfi}
    & \xmark
    & \textbf{38.32} & \textbf{0.9871} & 0.0151 & 0.0218
    & \underline{36.45} & 0.9811 & 0.0196 & 0.0343
    \\
    UPRNet-LARGE~\cite{uprnet}
    & \xmark
    & \underline{38.09} & \underline{0.9861} & 0.0150 & 0.0209
    & 36.42 & \underline{0.9815} & 0.0201 & 0.0342
    \\
    CAIN~\cite{cain}
    & \cmark
    & 35.11 & 0.9761 & 0.0254 & 0.0383
    & 34.65 & 0.9729 & 0.0306 & 0.0483
    \\
    FILM-${\mathcal{L}_{vgg}}$~\cite{film}
    & \cmark
    & 37.28 & 0.9843 & 0.0096 & 0.0148
    & 35.62 & 0.9784 & 0.0137 & 0.0229
    \\
    FILM-$\mathcal{L}_s$~\cite{film}
    & \cmark
    & 37.38 & 0.9844 & \textbf{0.0093} & \underline{0.0140}
    & 35.71 & 0.9787 & \textbf{0.0131} & \underline{0.0224}
    \\
    LDMVFI~\cite{LDMVFI}
    & \cmark
    & 34.03 & 0.9648 & 0.0195 & 0.0261
    & 33.09 & 0.9558 & 0.0233 & 0.0327
    \\
    PerVFI~\cite{pervfi}
    & \cmark
    & 35.00 & 0.9751 & 0.0142 & 0.0163
    & 34.00 & 0.9675 & 0.0179 & 0.0248
    \\
    MoMo (Ours)
    & \cmark
    & 36.77 & 0.9806 & \underline{0.0094} & \textbf{0.0126}
    & 34.94 & 0.9756 & \underline{0.0136} & \textbf{0.0203}
    \\
    MoMo-10M (Ours)
    & \cmark
    & 36.52 & 0.9801 & 0.0100 & 0.0139
    & 34.82 & 0.9752 & 0.0138 & 0.0206
    \\
    \bottomrule
    
    \end{tabular}
    \caption{Full quantitative results including the fidelity metrics (PSNR, SSIM) on Middlebury~\cite{middlebury} and Vimeo90k~\cite{toflow} benchmarks.
    The best results are in \textbf{bold}, and the second best is \underline{underlined}, respectively.
    }
    \label{tab:mb_vm_full}
\end{table*}

\begin{table*}[t!]
    \centering
    \setlength{\tabcolsep}{1mm}
    \begin{tabular}{l c cccc cccc}
    
    \toprule
    
    \multirowcell{2}{Method} & \multirowcell{2}{Perception-\\oriented loss} & \multicolumn{4}{c}{Xiph-2K} & \multicolumn{4}{c}{Xiph-4K} \\
    && PSNR & SSIM & LPIPS & DISTS & PSNR & SSIM & LPIPS & DISTS\\
    
    \midrule\midrule
    
    ABME~\cite{abme}
    & \xmark
    & 36.50 & 0.9668 & 0.1071 & 0.0581
    & 33.72 & 0.9452 & 0.2361 & 0.1108
    \\
    XVFI$_v$~\cite{xvfi}
    & \xmark
    & 35.17 & 0.9625 & 0.0844 & 0.0418
    & 32.45 & 0.9274 & 0.1835 & 0.0779
    \\
    IFRNet-Large~\cite{ifrnet}
    & \xmark
    & 36.40 & 0.9646 & 0.0681 & 0.0372
    & 33.71 & 0.9425 & 0.1364 & 0.0665
    \\
    RIFE~\cite{rife}
    & \xmark
    & 36.06 & 0.9642 & 0.0918 & 0.0481
    & 33.21 & 0.9413 & 0.2072 & 0.0915
    \\
    FILM-$\mathcal{L}_1$~\cite{film}
    & \xmark
    & 36.53 & 0.9663 & 0.0906 & 0.0510
    & 33.83 & 0.9439 & 0.1841 & 0.0884
    \\
    AMT-G~\cite{amt}  
    & \xmark
    & 36.29 & 0.9647 & 0.1061 & 0.0563
    & \underline{34.55} & \underline{0.9472} & 0.2054 & 0.1005
    \\
    EMA-VFI~\cite{emavfi}
    & \xmark
    & \underline{36.74} & \underline{0.9675} & 0.1024 & 0.0550
    & \underline{34.55} & \textbf{0.9486} & 0.2258 & 0.1049
    \\
    UPRNet-LARGE~\cite{uprnet}
    & \xmark
    & \textbf{37.13} & \textbf{0.9691} & 0.1010 & 0.0553
    & \textbf{34.57} & 0.9388 & 0.2150 & 0.1017
    \\
    CAIN~\cite{cain}
    & \cmark
    & 35.18 & 0.9625 & 0.1025 & 0.0533
    & 32.55 & 0.9398 & 0.2229 & 0.0980
    \\
    FILM-${\mathcal{L}_{vgg}}$~\cite{film}
    & \cmark
    & 36.29 & 0.9626 & 0.0355 & 0.0238
    & 33.44 & 0.9356 & 0.0754 & 0.0406
    \\
    FILM-$\mathcal{L}_s$~\cite{film}
    & \cmark
    & 36.30 & 0.9616 & \underline{0.0330} & 0.0237
    & 33.37 & 0.9323 & \underline{0.0703} & 0.0385
    \\
    LDMVFI~\cite{LDMVFI}
    & \cmark
    & 33.82 & 0.9494 & 0.0420 & 0.0163
    & 31.39 & 0.9214 & 0.0859 & 0.0359
    \\
    PerVFI~\cite{pervfi}
    & \cmark
    & 34.69 & 0.9541 & 0.0381 & \underline{0.0153}
    & 32.30 & 0.9149 & 0.0858 & \underline{0.0331}
    \\
    MoMo (Ours)
    & \cmark
    & 35.38 & 0.9553 & \textbf{0.0300} & \textbf{0.0119}
    & 33.09 & 0.9293 & \textbf{0.0631} & \textbf{0.0274}
    \\
    MoMo-10M (Ours)
    & \cmark
    & 35.23 & 0.9548 & 0.0303 & 0.0120
    & 32.97 & 0.9281 & 0.0638 & 0.0275
    \\
    \bottomrule
    
    \end{tabular}
    \caption{Full quantitative results including the fidelity metrics (PSNR, SSIM) on Xiph-2K and Xiph-4K~\cite{xiph,softsplat}.
    The best results are in \textbf{bold}, and the second best is \underline{underlined}, respectively.
    }
    \label{tab:xiph_full}
\end{table*}

\end{document}